\documentclass[10pt,conference]{IEEEtran}

\newif\ifieee
\ieeefalse

\newif\ifarxiv
\arxivtrue

\ifieee
    \ifarxiv
        \PackageError{Configuration}
        {IEEE and arXiv cannot both be enabled}
        {Set either \string\ieeefalse\space or \string\arxivfalse.}
    \fi
\fi

\usepackage{cite}

\ifCLASSINFOpdf
   \usepackage[pdftex]{graphicx}
   \DeclareGraphicsExtensions{.pdf,.jpeg,.png}
\else
   \usepackage[dvips]{graphicx}
   \DeclareGraphicsExtensions{.eps}
\fi

\usepackage[cmex10]{amsmath}
\usepackage[utf8]{inputenc}
\usepackage[T1]{fontenc}
\usepackage{newunicodechar}
\newunicodechar{→}{$\rightarrow$}

\usepackage{balance}

\usepackage{subcaption}
\usepackage{multirow}
\usepackage{booktabs}
\usepackage[dvipsnames]{xcolor}%
\usepackage{array}
\usepackage{url}
\usepackage{tikz}
\usepackage{ifthen}
\usepackage[acronym]{glossaries}
\glsdisablehyper

\newcommand\major[1]{#1} %

\ifieee
\else
    \usepackage[colorlinks=true,allcolors=black]{hyperref} %
\fi

\usepackage[capitalise]{cleveref}

\usepackage{transparent}
\usepackage{tikz}
\ifarxiv
    \newcommand\copyrighttext{%
      \scriptsize Accepted for presentation at SIBGRAPI 2026. The final published version will be available on IEEE~Xplore.}
    \newcommand\copyrightnotice{%
    \begin{tikzpicture}[remember picture,overlay]
    \node[anchor=south,yshift=30pt,xshift=0pt] at (current page.south) {\fbox{\transparent{0.85}\parbox{\dimexpr0.6\textwidth-\fboxsep-\fboxrule\relax}{\copyrighttext}}};
    \end{tikzpicture}%
    }
\else
\fi

\newif\iffinal
\finaltrue
\newcommand{\cmtid}{161}

\iffinal
\else
\usepackage[switch]{lineno}
\fi

\ifieee
\IEEEoverridecommandlockouts
\IEEEpubid{\makebox[\columnwidth]{979-8-3195-0255-1/26/\$31.00~\copyright2026 IEEE \hfill}
\hspace{\columnsep}\makebox[\columnwidth]{ }}
\else
\fi

\begin{document}

\title{On the Role of MRI Sequences in Cross-Dataset Generalization for Brain Tumor Segmentation}

\iffinal
\author{\IEEEauthorblockN{Henrique Zan Grande\IEEEauthorrefmark{1},
João G. Pitol\IEEEauthorrefmark{1},
Lucas B. Schuck\IEEEauthorrefmark{1}, Rafael V. Serenato\IEEEauthorrefmark{1},\\Rayson Laroca\IEEEauthorrefmark{1}, and Andre Gustavo Hochuli\IEEEauthorrefmark{1}}
\IEEEauthorblockA{
        \IEEEauthorrefmark{1}\hspace{0.15mm}Pontifical Catholic University of Paran\'a, Curitiba, Brazil\\[0.75ex]
            \IEEEauthorrefmark{1}\hspace{-0.35mm}\tt{\small{\{joao.pitol,lucas.schuck,rafael.serenato\}}@pucpr.edu.br} \\ \hspace{-1.5mm}\IEEEauthorrefmark{1}{\tt\small {\{henrique.zgrande,rayson,aghochuli\}}@ppgia.pucpr.br}}%
    }
\fi

\newcommand*{\RL}[2][]{\textcolor{Rhodamine}{[\textbf{\ifthenelse{\equal{#1}{}}{RL}{RL(#1)}}: #2]}}

\maketitle

\ifarxiv
    \copyrightnotice
\else
\fi

\iffinal
    \newcommand{\urlSupplementary}{\url{https://github.com/henrique-zan/brain_tumor_segmentation/}}
\fi

\begin{abstract}

Brain tumor segmentation in magnetic resonance imaging (MRI) is a critical task for diagnosis and treatment planning. Despite the success of deep learning architectures such as U-Net and its variants, performance degradation across datasets remains a major challenge, particularly under domain shift and limited annotated data. To address this issue, this study systematically evaluates how individual MRI sequences influence model robustness across two well-known datasets. A ResUNet-based framework is employed, where each modality is trained independently to isolate its effect under a controlled cross-dataset evaluation protocol with tumor size stratification, without target-domain training, or with limited domain adaptation. Results show that the T2f/FLAIR sequence achieves the best cross-dataset performance, with Dice scores exceeding 75\%. It consistently outperforms other modalities across most tumor size ranges, while multi-sequence training further improves performance. Additionally, even limited target-domain adaptation yields rapid initial gains, reducing the need for extensive annotations and costly retraining.
Our source code is publicly available at \textit{\urlSupplementary}.

\end{abstract}

\IEEEpeerreviewmaketitle

\section{Introduction}\label{sec:intro}

\glsresetall

Automatic brain tumor segmentation in magnetic resonance imaging (MRI) is a prominent application of deep learning in medical computer vision~\cite{ali2020brain, solanki2023brain, soomro2023image}.
This task plays a critical role in diagnosis, surgical planning, and clinical follow-up, supporting clinicians in analyzing imaging data.
Advances in convolutional neural network architectures, especially U-Net~\cite{10.1007/978-3-319-24574-4_28} and its variants, have achieved strong performance on benchmark datasets and competitions such as Brain Tumor Segmentation~(BraTS)\cite{brats2020,brats2024}, establishing them as a standard approach for brain lesion segmentation.

Despite these advances, segmentation models often fail to generalize across datasets~\cite{zhang2020generalizing}.
Differences in acquisition protocols, scanners, patient populations, and image characteristics create domain shifts that can substantially degrade performance on unseen data.
This cross-dataset generalization problem is further aggravated by the limited availability of annotated medical images, which restricts model robustness and real-world applicability~\cite{zhang2020generalizing}.
The choice of MRI sequence may be particularly important because T1n, T1c, T2w, and T2f/FLAIR provide complementary information about tissue contrast, edema, and tumor boundaries~\cite{yin2025tumor, xing20253d}.

Previous studies have primarily sought to maximize segmentation accuracy through architectural improvements or multimodal fusion~\cite{yin2025tumor, mangayarkarasi2024brain}.
However, fewer studies have systematically isolated the contribution of each MRI sequence under cross-dataset domain shift.
A modality-specific analysis can reveal which sequences yield more transferable representations and can guide model design when computational resources or MRI sequences are limited.
Moreover, evaluating adaptation with small amounts of labeled target-domain data can clarify the trade-off between annotation effort and performance improvement.

Based on this motivation, we investigate the following research questions:
\begin{itemize}
    \item(\textbf{RQ1}) How does segmentation performance degrade across domains as a function of MRI modality?

    \item(\textbf{RQ2}) Which MRI sequences exhibit greater robustness in cross-dataset binary brain tumor segmentation?

    \item(\textbf{RQ3}) What is the relationship between target-domain annotation effort and performance improvement for different MRI modalities?
\end{itemize}

To address these questions, we conduct a systematic comparative analysis of cross-dataset generalization across MRI sequences for binary brain tumor segmentation (tumor vs. non-tumor).
A standardized pipeline based on the widely used 2D ResUNet \cite{mangayarkarasi2024brain} is employed, where each modality is trained and evaluated independently under a controlled protocol~\cite{zhang2020generalizing} to isolate its contribution to robustness. Additionally, we apply fine-tuning to the best-performing modality to assess the impact of limited target-domain data on generalization, with emphasis on a controlled and analysis-driven evaluation~setting.

This work makes four main contributions: (i)~a controlled evaluation of modality-specific robustness under domain shift; (ii)~a quantitative assessment of the generalization gap across modalities; (iii)~the identification of the most robust MRI sequence for cross-dataset binary segmentation; and (iv)~an analysis of data-efficient domain adaptation, relating annotation budget to performance gains.

The remainder of this paper is organized as follows. \cref{sec:review} reviews related work and highlights the research gap. \cref{sec:problem} describes the problem, datasets, materials, and methodology. \cref{sec:exp_disc} presents the experimental setup, evaluation metrics and discusses the results. Finally, \cref{sec:conclusions} concludes the paper and outlines directions for future~work. 

\section{Literature Review}
\label{sec:review}

The literature on brain tumor segmentation in MRI is largely dominated by convolutional neural network-based approaches, particularly U-Net \cite{10.1007/978-3-319-24574-4_28} and its variants~\cite{ali2020brain, solanki2023brain}. Prior work has focused on architectural improvements and multimodal fusion strategies to enhance segmentation accuracy. Despite these advances, challenges related to domain shift and cross-dataset generalization remain insufficiently addressed.

U-Net has become a foundational architecture for brain tumor segmentation because its encoder-decoder structure and skip connections combine contextual representation with spatial detail~\cite{yang2018automatic}.
However, performance can deteriorate in heterogeneous imaging settings, where variations in tumor morphology and image intensity hinder generalization~\cite{goswami2020analysis}.

Recent variants such as ResUNet, ResUNet+, and nnU-Net address some of these limitations through residual blocks, attention mechanisms, and automated configuration~\cite{huang2024segmenting, mangayarkarasi2024brain, metlek2023resunet+}.
ResUNet combines the localization capabilities of U-Net with residual learning, which facilitates optimization and supports deeper feature extraction~\cite{huang2024segmenting}.
Mangayarkarasi et al.~\cite{mangayarkarasi2024brain} reported an accuracy of 93.75\% for ResUNet on the LGG MRI dataset.
Metlek \& Çetiner~\cite{metlek2023resunet+} reported that ResUNet+ improved the mean Dice score for whole-tumor segmentation by up to 5.58\% over the compared methods.
In parallel, nnU-Net has demonstrated strong and reproducible performance by automatically configuring architectural and preprocessing choices~\cite{huang2024segmenting}.

Multimodal fusion has also been extensively studied because T1, T1c, T2, and T2f/FLAIR provide complementary contrast and structural information.
Studies by Yin et al.~\cite{yin2025tumor} and Xing et al.~\cite{xing20253d} show that combining modalities can improve segmentation accuracy and robustness over single-sequence inputs, with reported Dice scores above 90\%.
Three-dimensional architectures, including ResUNet variants with dual or mirrored encoders, can better exploit volumetric continuity and inter-slice consistency.
However, they require more computation and may be more sensitive to class imbalance~\cite{xing20253d}.

Although these methods achieve strong in-domain results, their robustness across heterogeneous datasets remains limited.
Differences in acquisition protocols, scanners, and intensity distributions can induce substantial performance degradation on unseen domains~\cite{zhang2020generalizing, metlek2023resunet+}.
Multimodal inputs may reduce this sensitivity, but the individual contribution of each MRI sequence to cross-domain robustness has not been systematically quantified, especially under limited annotation budgets.
This study addresses that gap through a controlled modality-specific evaluation and a data-efficient domain-adaptation analysis.

\section{Problem Statement}
\label{sec:problem}

Consider a model trained on one BraTS cohort and deployed on another using a single MRI sequence.
The central problem is to determine how much segmentation performance is lost during this transfer, whether some sequences preserve accuracy better than others, and how efficiently a small labeled target subset can compensate for the shift.

To isolate these factors, we evaluate every modality with the same 2D ResUNet architecture, data preparation procedure, and optimization settings.
The comparison covers in-domain evaluation, direct cross-dataset transfer, joint training on both datasets, and incremental target-domain fine-tuning.

The following section details the experimental design and methodological choices adopted to ensure a controlled and reproducible evaluation. We first describe the datasets and their characteristics, followed by the partitioning and preprocessing pipeline applied uniformly across all experiments. Next, the 2D ResUNet architecture and training configuration are presented, including loss functions and optimization strategy.
Finally, we outline the evaluation.

\subsection{Datasets}\label{sec:datasets}

\begin{figure}[!htb]
    \centering
    \captionsetup[subfigure]{
        labelformat=empty
    }

    \begin{subfigure}{0.09\textwidth}
        \includegraphics[width=\linewidth, angle=-90, trim=20pt 20pt 20pt 20pt, clip]{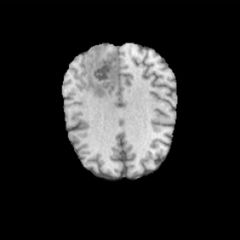}
    \end{subfigure}
    \begin{subfigure}{0.09\textwidth}
        \includegraphics[width=\linewidth, angle=-90, trim=20pt 20pt 20pt 20pt, clip]{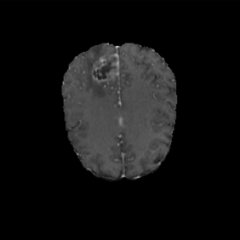}
    \end{subfigure}
    \begin{subfigure}{0.09\textwidth}
        \includegraphics[width=\linewidth, angle=-90, trim=20pt 20pt 20pt 20pt, clip]{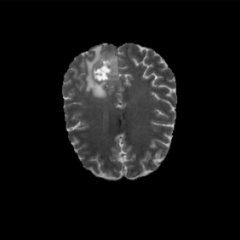}
    \end{subfigure}
    \begin{subfigure}{0.09\textwidth}
        \includegraphics[width=\linewidth, angle=-90, trim=20pt 20pt 20pt 20pt, clip]{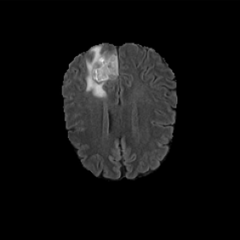}
    \end{subfigure}
    \begin{subfigure}{0.09\textwidth}
        \includegraphics[width=\linewidth, angle=-90, trim=20pt 20pt 20pt 20pt, clip]{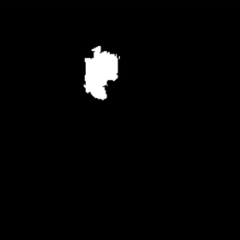}
    \end{subfigure}

    \vspace{0.3em}

    \begin{subfigure}{0.09\textwidth}
        \includegraphics[width=\linewidth, angle=-90, trim=20pt 20pt 20pt 20pt, clip]{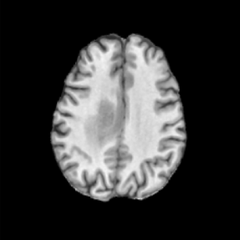}
    \end{subfigure}
    \begin{subfigure}{0.09\textwidth}
        \includegraphics[width=\linewidth, angle=-90, trim=20pt 20pt 20pt 20pt, clip]{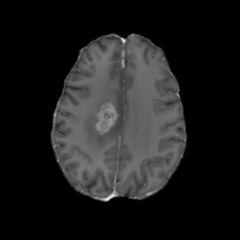}
    \end{subfigure}
    \begin{subfigure}{0.09\textwidth}
        \includegraphics[width=\linewidth, angle=-90, trim=20pt 20pt 20pt 20pt, clip]{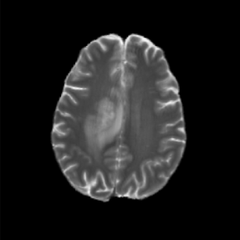}
    \end{subfigure}
    \begin{subfigure}{0.09\textwidth}
        \includegraphics[width=\linewidth, angle=-90, trim=20pt 20pt 20pt 20pt, clip]{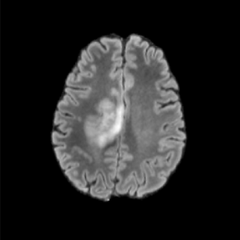}
    \end{subfigure}
    \begin{subfigure}{0.09\textwidth}
        \includegraphics[width=\linewidth, angle=-90, trim=20pt 20pt 20pt 20pt, clip]{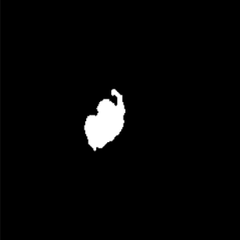}
    \end{subfigure}

    \vspace{-0.2em}

    \begin{subfigure}{0.09\textwidth}
        \caption{T1n}
    \end{subfigure}
    \begin{subfigure}{0.09\textwidth}
        \caption{T1c}
    \end{subfigure}
    \begin{subfigure}{0.09\textwidth}
        \caption{T2w}
    \end{subfigure}
    \begin{subfigure}{0.09\textwidth}
        \caption{T2f/Flair}
    \end{subfigure}
    \begin{subfigure}{0.09\textwidth}
        \caption{Mask}
    \end{subfigure}

    \caption{\major{Representative cases from BraTS 2020~(top) and BraTS 2024~(bottom) across the four MRI sequences and the corresponding binary mask.}}
    \label{fig:brats_comparison}
\end{figure}

The \textit{Brain Tumor Segmentation}~(BraTS) challenge datasets from 2020~\cite{brats2020} and 2024 (Adult glioma segmentation task)~\cite{brats2024} were selected due to their widespread adoption as standardized benchmarks and their complementary characteristics for evaluating model robustness. BraTS 2020~\cite{brats2020} comprises 369 multimodal MRI studies and serves as a well-established, controlled benchmark for method comparison. In contrast, BraTS 2024~\cite{brats2024} extends this scale to a substantially larger and more heterogeneous cohort, totaling approximately 1{,}350 cases, reflecting recent advances in data collection and acquisition diversity. Both datasets provide four MRI sequences (T1n, T1c, T2w, and T2f/FLAIR) along with expert-annotated segmentation masks for tumor subregions, \major{as shown in \cref{fig:brats_comparison}}.

Although the datasets follow standardized annotation and preprocessing practices, they differ in patient characteristics, tumor morphology, scanners, and acquisition protocols.
\major{For example, BraTS 2020 comprises pre-operative adult glioma scans from a limited set of contributing institutions, whereas BraTS 2024 is substantially larger and more heterogeneous, aggregating post-treatment acquisitions from a broader range of sites and scanners.}
These differences create a realistic domain shift while preserving a common segmentation task. The two datasets are therefore suitable for evaluating cross-dataset generalization and modality-specific robustness. Whenever available, we retained the original patient-level organization and naming conventions to support~reproducibility.

Data preparation followed a patient-level holdout strategy, with 70\% of the patients assigned to training, 15\% to validation, and 15\% to testing.
This partitioning prevents data leakage by ensuring that slices from the same patient do not appear in multiple subsets.
All images were resized to $240 \times 240$ pixels using the same preprocessing pipeline for both datasets.
For each MRI modality, we extracted 2D slices containing tumor tissue to focus training and evaluation on informative regions and reduce computational cost.
The original multiclass annotations were converted into binary masks representing tumor and non-tumor regions.

Data augmentation was applied only to the training set and included $90^\circ$ rotations and horizontal and vertical flips.
We deliberately excluded more aggressive intensity and geometric transformations because they may alter anatomical structures or create unrealistic tumor appearances~\cite{zhang2020generalizing,krinski2023dacov}.

\subsection{Architecture}\label{sec:architecture}

\begin{figure*}[!t]
  \centering
  \includegraphics[width=1\textwidth]{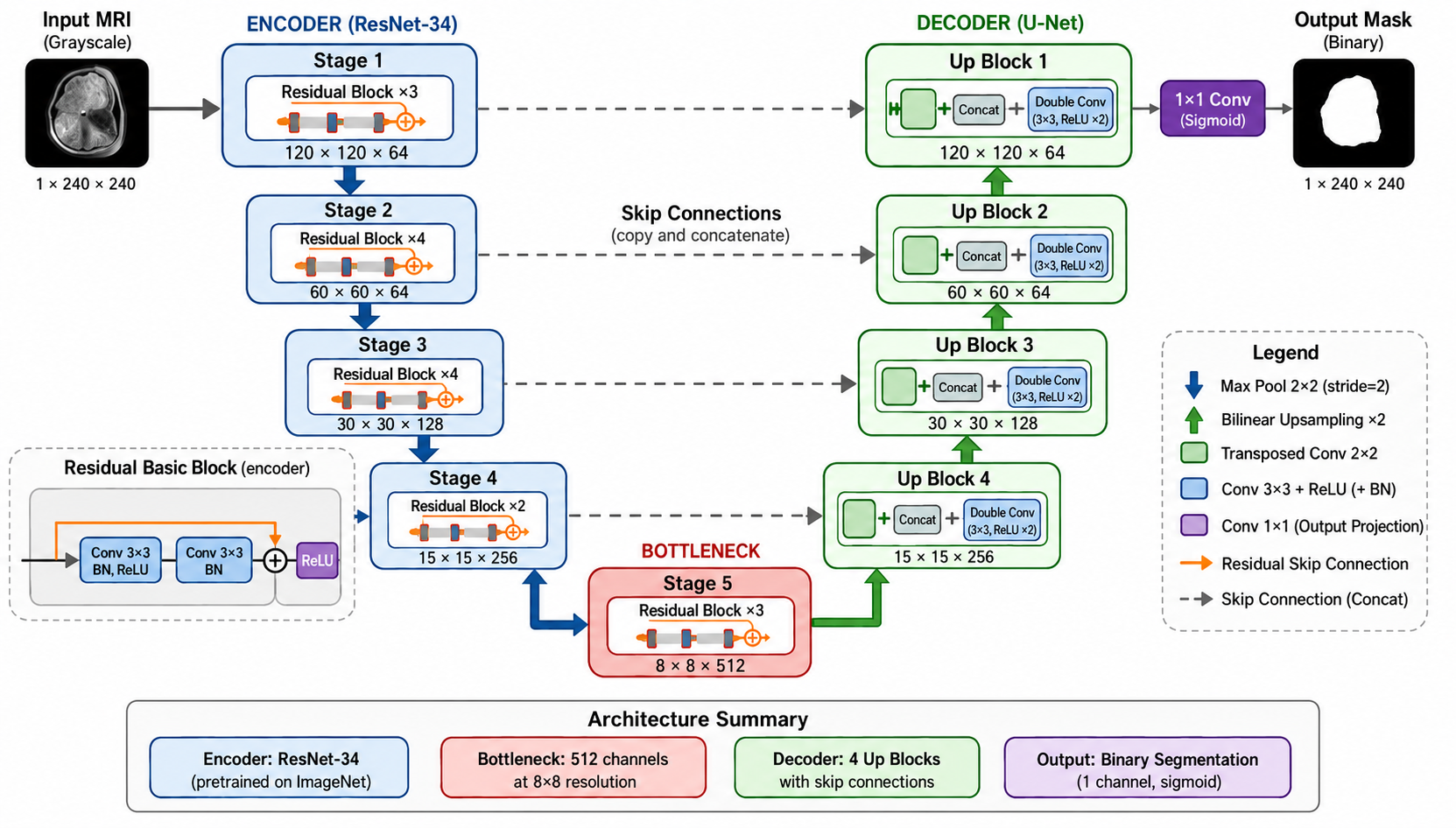}

  \vspace{-1mm}
  
  \caption{Overview of the implemented 2D ResUNet architecture.}
  \label{img:arquitetura}
\end{figure*}

The adopted architecture is a 2D ResUNet based on a ResNet-34 encoder~\cite{mangayarkarasi2024brain} (see \cref{img:arquitetura}).
ResUNet extends U-Net with residual learning to improve optimization, feature representation, and training stability~\cite{huang2024segmenting, mangayarkarasi2024brain, metlek2023resunet+}.
\major{The encoder uses a pretrained ResNet-34 backbone with channel widths of 64, 128, 256, and 512 to extract hierarchical features.
Residual identity mappings facilitate gradient propagation, while the encoder progressively captures multi-scale contextual information.
Skip connections transfer fine-grained spatial features from each encoder stage to the corresponding decoder~stage.}

\major{The decoder contains four upsampling stages with channel progression $512 \rightarrow 256 \rightarrow 128 \rightarrow 64 \rightarrow 64$.
Each stage combines transposed convolution, feature concatenation, and two $3 \times 3$ Conv-BatchNorm-ReLU blocks for feature fusion and spatial refinement.
Dropout with $p=0.2$ is applied within the convolutional blocks to reduce overfitting, particularly in limited-data and cross-dataset settings.
The network receives a single-channel MRI slice and produces a binary segmentation map through a final $1 \times 1$ convolution.}

\textbf{Training Protocol and Hyperparameters}. \major{Models were trained independently for each MRI modality using a supervised objective that combines Dice loss and binary cross-entropy~\cite{zhang2020generalizing, xing20253d}.
Training was conducted for up to 80 epochs with a batch size of 32 and an initial learning rate of $1 \times 10^{-3}$.
The learning rate was reduced adaptively according to validation performance.
Early stopping with a patience of eight epochs was based on the validation loss.
For each experiment, we retained the checkpoint with the lowest validation loss\footnote{To ensure full reproducibility, all models, trained weights, source code, and experimental artifacts are publicly available at \url{https://github.com/henrique-zan/brain_tumor_segmentation/}.}.}

\section{Experiments and Discussion}\label{sec:exp_disc}

The experimental analysis comprises three parts using BraTS 2020~\cite{brats2020} and BraTS 2024~\cite{brats2024}.
First, \cref{sec:generalization} compares in-domain, cross-dataset, and multi-domain protocols to answer RQ1 and RQ2.
Second, \cref{sec:domain_adapt} measures how increasing proportions of labeled target-domain data affect fine-tuning performance, addressing RQ3.
Third, \cref{sec:qualitative} examines tumor-size quartiles and representative segmentation cases.
The size-stratified analysis accounts for the greater sensitivity of overlap-based metrics such as Dice to small absolute errors in small lesions.

\major{Performance was measured using the Dice coefficient and Intersection over Union~(IoU), two overlap-based metrics widely adopted in medical image segmentation~\cite{rajendran2023automated, metlek2023resunet+}.
Both metrics quantify agreement between the predicted and reference masks.
Metrics were computed over all tumor-containing test slices and averaged within each test set to enable consistent comparisons across modalities and experimental~settings.}

\subsection{Generalization Evaluation (RQ1/RQ2)}\label{sec:generalization}

To rigorously assess cross-domain robustness, we design evaluation protocols that progressively increase the distributional discrepancy between training and test data, enabling a controlled analysis of generalization behavior across MRI~modalities.

\textbf{Baseline}. In the baseline setting, models are trained and evaluated within the same domain (BraTS 2020 or BraTS 2024 independently), providing an upper-bound reference under in-distribution conditions.
\major{As training, validation, and test sets share the same underlying data distribution, this scenario isolates the intrinsic segmentation capability of each modality without domain shift effects.}

\textbf{Cross-Dataset}. To explicitly evaluate generalization under domain shift, models trained on one dataset are evaluated directly on the other without adaptation. Two configurations are considered: BraTS 2024$\rightarrow$2020 and BraTS 2020$\rightarrow$2024. This protocol introduces discrepancies in acquisition protocols, intensity distributions, and population characteristics, providing a realistic assessment of performance degradation across domains. It enables quantifying modality-specific robustness and identifying which MRI sequences are inherently more resilient to distributional shifts~\cite{zhang2020generalizing}.

\textbf{Multi-Domain Training}. In this setting, training data from BraTS 2020 and BraTS 2024 are jointly used in balanced proportions (50/50), while preserving patient-level holdout splits to avoid data leakage. This protocol investigates whether exposure to heterogeneous data distributions improves generalization by promoting the learning of domain-invariant representations, and assesses the extent to which shared features across domains can mitigate performance degradation.

\major{The results summarized in Table~\ref{tab:resultados}} provide key insights into cross-dataset generalization across MRI modalities. A consistent gap between in-domain and cross-domain performance is observed, confirming the sensitivity of segmentation models to domain shift~(RQ1). While baseline results establish an upper bound, with FLAIR (T2F) and T2W achieving the highest Dice scores (e.g., 77.3--78.3 and 73.3--74.2), cross-dataset evaluation reveals substantial performance degradation, particularly for T2F (77.3$\rightarrow$58.2) and T2W (73.3$\rightarrow$62.3). Despite this, T2F maintains the highest cross-domain performance (up to 76.2), followed by T2W, whereas T1-based modalities show lower robustness (e.g., $\approx$49.7 Dice).

\begin{table}[!htb]
\centering
\renewcommand{\arraystretch}{1.2}
\setlength{\tabcolsep}{3pt}
\caption{\major{Segmentation performance (\%) across evaluation protocols and MRI modalities.}}
\resizebox{\columnwidth}{!}{
\begin{tabular}{lcc|cc|cc|cc}
\toprule
\multirow{2}{*}{\textbf{Scenario}} &
\multicolumn{2}{c|}{\textbf{T1C}} &
\multicolumn{2}{c|}{\textbf{T1N}} &
\multicolumn{2}{c|}{\textbf{T2W}} &
\multicolumn{2}{c}{\textbf{T2F}} \\
\cmidrule(lr){2-3}
\cmidrule(lr){4-5}
\cmidrule(lr){6-7}
\cmidrule(l){8-9}
& \textbf{\major{Dice}} & \textbf{\major{IoU}}
& \textbf{\major{Dice}} & \textbf{\major{IoU}}
& \textbf{\major{Dice}} & \textbf{\major{IoU}}
& \textbf{\major{Dice}} & \textbf{\major{IoU}} \\
\midrule
Baseline 2020        & 61.3 & 51.2 & 62.9 & 52.9 & 73.3 & 64.2 & \textbf{77.3} & \textbf{68.4} \\
Baseline 2024        & 66.1 & 54.7 & 67.3 & 56.0 & 74.2 & 63.5 & \textbf{78.3} & \textbf{68.6} \\
Cross-Dataset 20/24  & 49.7 & 38.8 & 49.7 & 39.1 & \textbf{62.3} & \textbf{51.4} & 58.2 & 46.9 \\
Cross-Dataset 24/20  & 65.3 & 55.5 & 58.3 & 48.8 & 70.4 & 61.2 & \textbf{76.2} & \textbf{67.6} \\
Multi-Domain 20/24   & 63.5 & 52.3 & 63.9 & 52.9 & 72.1 & 61.7 & \textbf{75.0} & \textbf{65.2} \\
\bottomrule
\end{tabular}
}
\label{tab:resultados}
\end{table}
From a representation learning perspective, T2-based modalities enable more transferable features, while T1-based inputs induce more domain-dependent representations. The smaller drop (e.g., T2F: 78.3$\rightarrow$76.2) suggests that training on BraTS 2024 yields more generalizable representations, driven by its larger scale and greater data diversity. Multi-domain training further mitigates degradation (e.g., T2F: 75.0 vs. 77.3/78.3), indicating that exposure to heterogeneous data promotes domain-invariant features. Overall, modality choice plays a central role in model generalization, with FLAIR providing the most robust representations under domain shift~(RQ2).

\subsection{Domain Adaptation (RQ3)}\label{sec:domain_adapt}

\begin{figure}[!htb]
    \centering
    \includegraphics[width=1\linewidth]{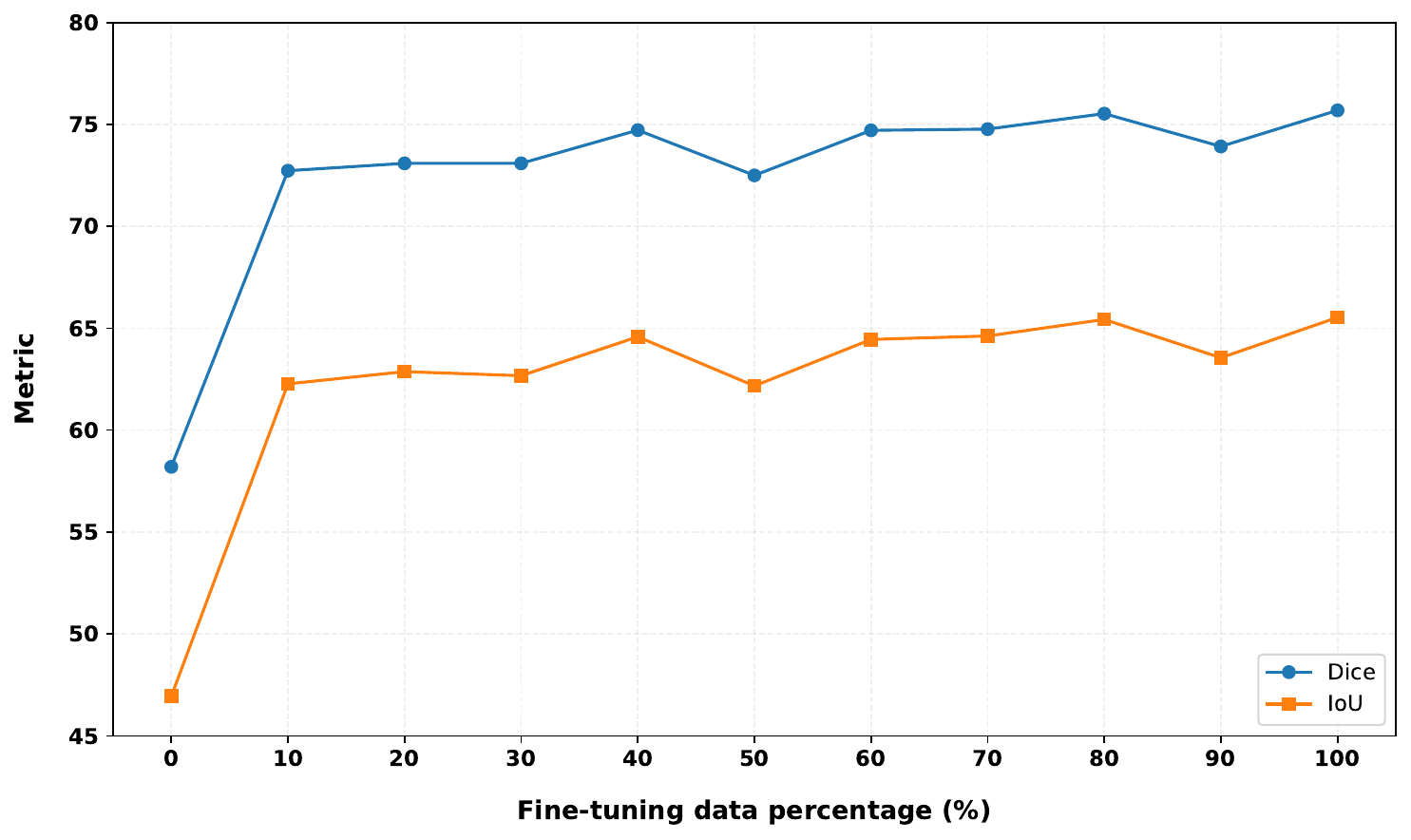}
    \caption{Domain adaptation performance of a model trained on BraTS 2020 and progressively fine-tuned with increasing fractions of BraTS 2024 data.}
    \label{fig:finetune_2020_2024}
\end{figure}

\begin{figure}[!htb]
    \centering
    \includegraphics[width=1\linewidth]{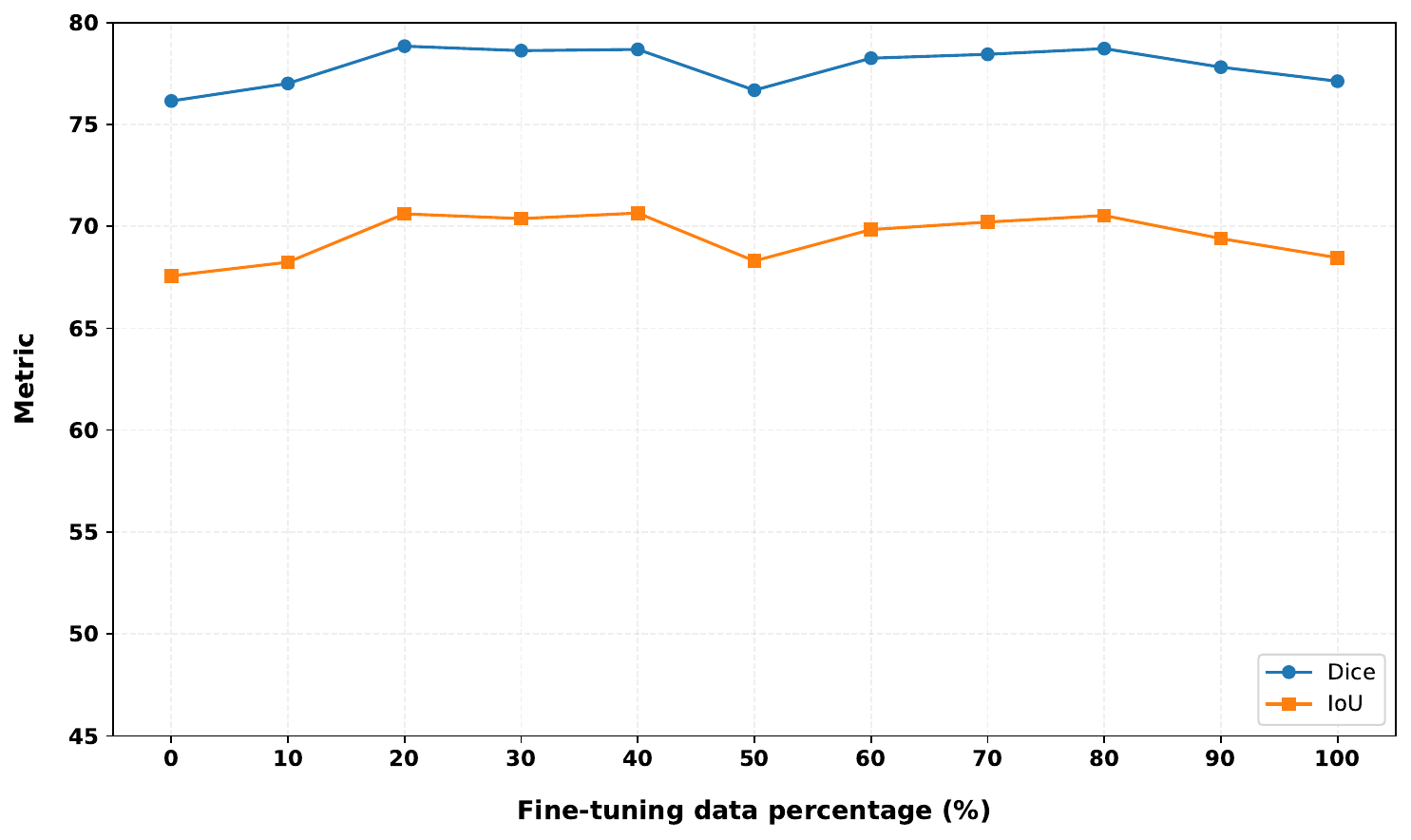}
    \caption{Domain adaptation performance of a model trained on BraTS 2024 and progressively fine-tuned with increasing fractions of BraTS 2020 data.}
    \label{fig:finetune_2024_2020}
\end{figure}

This scenario evaluates domain adaptation under limited labeled target-domain data by fine-tuning models, initially trained on a source dataset, using incremental fractions~(10\% steps) of annotated samples from the best-performing modality~(T2F/FLAIR). Results show that fine-tuning with only 10--20\% of target-domain data already approaches the upper-bound performance, after which a plateau is observed, indicating diminishing returns. This directly answers RQ3, demonstrating that a small amount of target data is sufficient to substantially improve generalization while reducing annotation and computational costs. Notably, the largest gains are observed when adapting models trained on BraTS 2020 to BraTS 2024 (see \cref{fig:finetune_2020_2024}), whereas the reverse direction (see \cref{fig:finetune_2024_2020}) yields more modest improvements ($\approx 5\%$ Dice), consistent with the generalization analysis in Section~\ref{sec:generalization}, which suggests that models trained on BraTS 2024 already learn more transferable representations.

\subsection{Qualitative Analysis}\label{sec:qualitative}

\major{\cref{tab:resultados_quartil_seq}} shows segmentation performance stratified by tumor size quartile. \major{Tumor size is defined as the number of foreground pixels in the ground-truth mask of each 2D slice. The slices are then divided into four equally populated quartiles, ranging from the smallest tumors~(Q1) to the largest~(Q4).} Dice and IoU increase consistently from Q1 to Q4. This trend reflects both the greater sensitivity of overlap-based metrics to segmentation errors in small structures and the richer morphological information available for larger tumors.

\begin{table}[!htb]
\centering
\renewcommand{\arraystretch}{1.2}
\setlength{\tabcolsep}{4pt}
\caption{\major{Segmentation performance (\%) stratified by tumor size quartile. Values in parentheses indicate the mean tumor area within each quartile.}}
\resizebox{\columnwidth}{!}{
\begin{tabular}{lcc|cc|cc|cc}
\toprule
\multirow{2}{*}{\textbf{Quartile / Size}} &
\multicolumn{2}{c|}{\textbf{T1C}} &
\multicolumn{2}{c|}{\textbf{T1N}} &
\multicolumn{2}{c|}{\textbf{T2W}} &
\multicolumn{2}{c}{\textbf{T2F}} \\
\cmidrule(lr){2-3}
\cmidrule(lr){4-5}
\cmidrule(lr){6-7}
\cmidrule(l){8-9}
& \textbf{\major{Dice}} & \textbf{\major{IoU}}
& \textbf{\major{Dice}} & \textbf{\major{IoU}}
& \textbf{\major{Dice}} & \textbf{\major{IoU}}
& \textbf{\major{Dice}} & \textbf{\major{IoU}} \\
\midrule
Q1 (171 px)  & 26.9 & 19.6 & 26.0 & 18.8 & 39.5 & 30.3 & \textbf{49.0} & \textbf{39.2} \\
Q2 (729 px)  & 63.3 & 51.1 & 64.0 & 51.4 & 75.3 & 63.8 & \textbf{77.6} & \textbf{66.8} \\
Q3 (1475 px) & 75.8 & 64.5 & 76.6 & 65.0 & \textbf{84.0} & 74.1 & \textbf{84.0} & \textbf{75.0} \\
Q4 (2957 px) & 79.7 & 68.6 & 81.3 & 70.1 & \textbf{85.7} & \textbf{76.4} & 84.8 & 76.0 \\
\bottomrule
\end{tabular}
}
\label{tab:resultados_quartil_seq}
\end{table}

Modality performance also varies with size. For small tumors~(Q1), T2f/FLAIR provides the best results, indicating higher sensitivity to subtle boundaries. For medium sizes~(Q2–Q3), T2f/FLAIR and T2w achieve comparable performance, both outperforming T1-based modalities. For large tumors~(Q4), T2w slightly surpasses T2f/FLAIR. Overall, T2f/FLAIR is preferable for small lesions, while T2f/FLAIR and T2w offer similar performance for larger volumes, with a slight advantage for T2w in the upper~quartile.

\begin{figure}[h!]
    \centering
    \begin{subfigure}{0.11\textwidth}
        \includegraphics[
            width=\linewidth,
            angle=-90,
            trim=20pt 20pt 20pt 20pt,
            clip
        ]{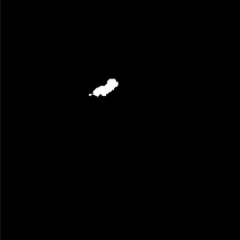}
        \vspace{-1mm}
        \caption{Q1}
    \end{subfigure}
    \begin{subfigure}{0.11\textwidth}
        \includegraphics[
            width=\linewidth,
            angle=-90,
            trim=20pt 20pt 20pt 20pt,
            clip
        ]{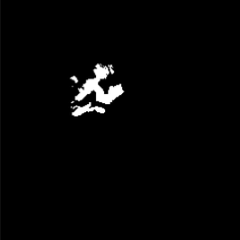}
        \vspace{-1mm}
        \caption{Q2}
    \end{subfigure}
    \begin{subfigure}{0.11\textwidth}
        \includegraphics[
            width=\linewidth,
            angle=-90,
            trim=20pt 20pt 20pt 20pt,
            clip
        ]{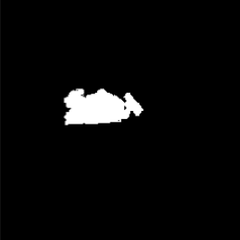}
        \vspace{-1mm}
        \caption{Q3}
    \end{subfigure}
    \begin{subfigure}{0.11\textwidth}
        \includegraphics[
            width=\linewidth,
            angle=-90,
            trim=20pt 20pt 20pt 20pt,
            clip
        ]{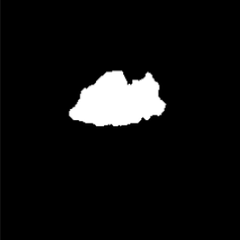}
        \vspace{-1mm}
        \caption{Q4}
    \end{subfigure}
    \caption{Representative masks for each tumor-size quartile.}
    \label{img:qurtil}
\end{figure}

To complement Table~\ref{tab:resultados_quartil_seq} and better illustrate performance across tumor sizes, \cref{img:qurtil} presents segmentation masks for the case closest to the mean tumor size of each quartile.

\begin{figure}[h!]
    \centering

    \begin{subfigure}{0.12\textwidth}
        \includegraphics[width=\linewidth, angle=-90, trim=20pt 20pt 20pt 20pt, clip]{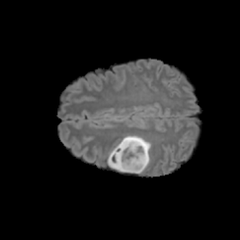}
    \end{subfigure}
    \begin{subfigure}{0.12\textwidth}
        \includegraphics[width=\linewidth, angle=-90, trim=20pt 20pt 20pt 20pt, clip]{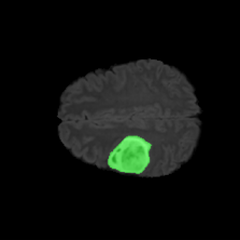}
    \end{subfigure}
    \begin{subfigure}{0.12\textwidth}
        \includegraphics[width=\linewidth, angle=-90, trim=20pt 20pt 20pt 20pt, clip]{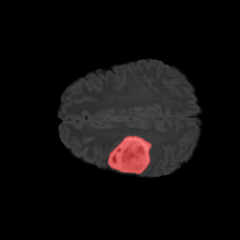}
    \end{subfigure}

    \vspace{0.3em}

    \begin{subfigure}{0.12\textwidth}
        \includegraphics[width=\linewidth, angle=-90, trim=10pt 20pt 30pt 20pt, clip]{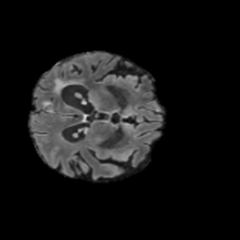}
    \end{subfigure}
    \begin{subfigure}{0.12\textwidth}
        \includegraphics[width=\linewidth, angle=-90, trim=10pt 20pt 30pt 20pt, clip]{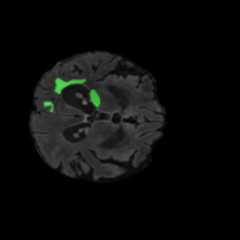}
    \end{subfigure}
    \begin{subfigure}{0.12\textwidth}
        \includegraphics[width=\linewidth, angle=-90, trim=10pt 20pt 30pt 20pt, clip]{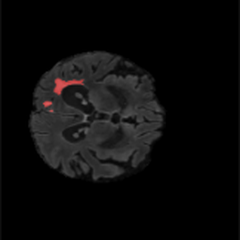}
    \end{subfigure}

    \vspace{0.3em}

    \begin{subfigure}{0.12\textwidth}
        \includegraphics[width=\linewidth, angle=-90, trim=30pt 20pt 10pt 20pt, clip]{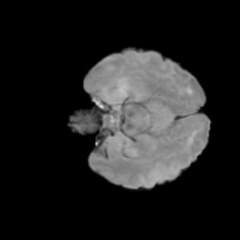}
    \end{subfigure}
    \begin{subfigure}{0.12\textwidth}
        \includegraphics[width=\linewidth, angle=-90, trim=30pt 20pt 10pt 20pt, clip]{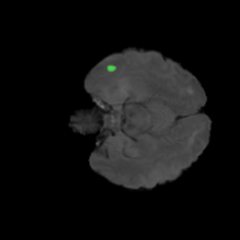}
    \end{subfigure}
    \begin{subfigure}{0.12\textwidth}
        \includegraphics[width=\linewidth, angle=-90, trim=30pt 20pt 10pt 20pt, clip]{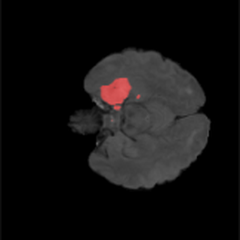}
    \end{subfigure}

    \vspace{0.2em}

    \begin{subfigure}{0.12\textwidth}
        \vspace{-1.5mm}
        \caption{Input}
    \end{subfigure}
    \begin{subfigure}{0.12\textwidth}
        \vspace{-1.5mm}
        \caption{GT}
    \end{subfigure}
    \begin{subfigure}{0.12\textwidth}
        \vspace{-1.5mm}
        \caption{Pred}
    \end{subfigure}

    \caption{\major{Segmentation examples: high-accuracy case (top, Dice:~99.1\%), intermediate case (middle, Dice:~70.1\%), and failure case (bottom, Dice:~0.0\%).}}
    \label{img:segmentation_examples}
\end{figure}

To contextualize the quantitative results, we present qualitative examples on the T2f/FLAIR sequence, including correct, intermediate, and failure cases~\major{(see \cref{img:segmentation_examples}).}
Each row shows the input image, the ground-truth overlay in green, and the predicted overlay in red.

\section{Conclusions}
\label{sec:conclusions}

This work investigated the role of MRI modalities in cross-dataset generalization through a standardized ResUNet-based pipeline and controlled evaluation protocols. The results support four main conclusions. First, T2f/FLAIR consistently provides the best performance and strongest generalization, followed closely by T2w, while T1-based modalities exhibit lower robustness under domain shift (RQ1/RQ2). Second, multi-domain training reduces the generalization gap, improving transfer performance and bringing results closer to in-domain conditions. Third, tumor size significantly affects performance: T2f/FLAIR is more effective for small lesions, whereas T2w becomes comparable or slightly superior for larger tumors. Fourth, domain adaptation via fine-tuning is highly data-efficient, with 10--20\% of target-domain data already approaching upper-bound performance and limited gains thereafter~(RQ3).

Overall, these findings demonstrate that model generalization is strongly influenced by modality choice, data diversity, and lesion characteristics, providing practical guidance for resource-constrained scenarios and modality selection in real-world applications.

\major{The insights gained from this work motivate several future research directions, including investigating improved embedding spaces and adaptive architectures, extending the analysis to multimodal and 3D models for enhanced volumetric consistency, validating the findings on independent clinical datasets beyond BraTS, developing more effective strategies for small lesion detection, and assessing foundation models with respect to generalization, sensitivity, and computational efficiency.}

\section*{Acknowledgments}

\iffinal    
    The authors acknowledge the financial support of the \textit{Coordenação de Aperfeiçoamento de Pessoal de Nível Superior~(CAPES)} --  Finance Code 001, the \textit{Financiadora de Estudos e Projetos}~(FINEP) for the High-Performance Computing (HPC) infrastructure provided by the \textit{Centro Integrado de Soluções em Inteligência Artificial}~(CISIA) at the \textit{Pontifícia Universidade Católica do Paraná}~(PUCPR), and the \textit{Fundação Araucária}, in partnership with the \textit{Secretaria da Ciência, Tecnologia e Ensino Superior do Estado do Paraná}~(SETI-PR), under Grant No. 653/2025~(FA/UNIVERSAL).
\fi

\balance

\bibliographystyle{IEEEtran}
\bibliography{bibtex}

\end{document}